\documentclass[11pt,letterpaper]{article}
\usepackage[T1]{fontenc}
\usepackage{ijcnlp2017}
\usepackage{times}
\usepackage{latexsym}

\usepackage{graphicx}
\usepackage{amsmath, amssymb}
\usepackage{bm}

\usepackage{fancyvrb}
\VerbatimFootnotes

\usepackage[noend]{algpseudocode}
\newcommand{\cev}[1]{\reflectbox{\ensuremath{\vec{\reflectbox{\ensuremath{#1}}}}}}

\DeclareMathOperator*{\RNN}{\overrightarrow{\mathrm{RNN}}}
\DeclareMathOperator*{\RNNL}{\overleftarrow{\mathrm{RNN}}}

\ijcnlpfinalcopy

\title{A Neural Language Model for Dynamically Representing\\the Meanings of Unknown Words and Entities in a Discourse}

\author{Sosuke Kobayashi \\ Preferred Networks, Inc., Japan \\ {\tt sosk@preferred.jp}
        \AND
        Naoaki Okazaki \\ Tokyo Institute of Technology, Japan \\ {\tt okazaki@c.titech.ac.jp} \And
        Kentaro Inui \\ Tohoku University / RIKEN, Japan \\ {\tt inui@ecei.tohoku.ac.jp}}

\date{}

\begin{document}

\maketitle

\begin{abstract}
This study addresses the
problem of identifying the meaning of unknown words or entities in a discourse
with respect to the word embedding approaches used in
neural language models.
We proposed a method for on-the-fly construction and exploitation
of word embeddings in both the input and output layers
of a neural model by tracking contexts.
This extends the dynamic entity representation used in \citet{Kobayashi16}
and incorporates a copy mechanism proposed independently by \citet{Gu16} and \citet{Gulcehre16}.
In addition, we construct
a new task and dataset called
{\it Anonymized Language Modeling}
for evaluating the ability to capture word meanings while reading.
Experiments conducted using our novel dataset
show that the proposed variant of RNN language model
outperformed the baseline model.
Furthermore, the experiments also demonstrate that
dynamic updates of an output layer
help a model predict reappearing entities,
whereas those of an input layer
are effective to predict words following reappearing entities.
\end{abstract}

\section{Introduction}
Language models that use probability distributions over sequences of words
are found in many natural language processing applications, including
speech recognition, machine translation, text summarization, and dialogue utterance generation.
Recent studies have demonstrated that language models trained using
neural network~\citep{Bengio03, Mikolov10} such as
recurrent neural network (RNN)~\citep{Jozefowicz16} and convolutional neural network~\citep{Yann16} achieve the best performance
across a range of corpora~\citep{Mikolov10, Chelba14, Merity16, Grave16}.

\begin{figure}[!t]
  \begin{center}
    \includegraphics[width=7.6cm,clip]{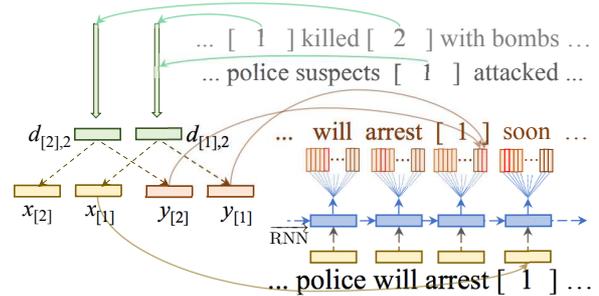}
    \caption{
Dynamic Neural Text Modeling: the embeddings of unknown words, denoted by coreference indexes ``[ k ]''  are dynamically computed and used in both the input and output layers ($x_{[\textrm{k}]}$ and $y_{[\textrm{k}]}$)
of a RNN language model.
These are constructed from contextual information ($d_{[\textrm{k}],\textrm{i}}$) preceding the current $(i+1)$-th sentence.}
    \label{fig:dynamic_lm}
  \end{center}
\end{figure}

However, current neural language models have a major drawback:
the language model works only when applied to a closed vocabulary of fixed size (usually comprising high-frequency words from the given training corpus).
All occurrences of out-of-vocabulary words are replaced with a single dummy token ``\verb|<|unk\verb|>|'', showing that the word is unknown.
For example, the word sequence, \textit{Pikotaro sings PPAP on YouTube} is treated as {\itshape \verb|<|unk\verb|>| sings \verb|<|unk\verb|>| on \verb|<|unk\verb|>|} assuming that the words \textit{Pikotaro}, \textit{PPAP}, and \textit{YouTube} are out of the vocabulary. The model therefore assumes that these words have the same meaning, which is clearly incorrect.
The derivation of \textit{meanings of unknown words} remains a persistent and nontrivial challenge when using word embeddings.

\begin{figure*}[!t]
  \begin{center}
    \includegraphics[width=14.5cm]{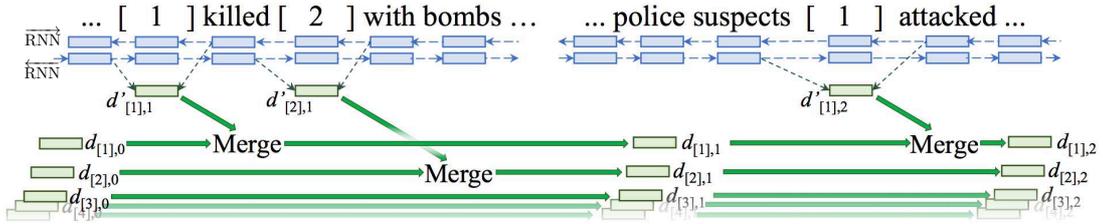}
    \caption{
Dynamic Neural Text Modeling: the meaning representation of each unknown word, denoted by a coreference index ``[ k ]'', is inferred from the local contexts in which it occurs.}
    \label{fig:context_merge}
  \end{center}
\end{figure*}

In addition, existing language models further assume that the meaning of a word is the same and universal across different documents. Neural language models also make this assumption and represent all occurrences of a word with a single word vector across all documents. However, the assumption of a universal meaning is also unlikely correct.
For example, the name \textit{John} is likely to refer to different individuals in different documents. In one story, \textit{John} may be a pianist while another \textit{John} denoted in a second story may be an infant. A model that represents all occurrences of \textit{John} with the same vector fails to capture the very different behavior expected from \textit{John} as a pianist and \textit{John} as an infant.

In this study, we address these issues and propose a novel neural language model that can build and dynamically change distributed representations of words based on the multi-sentential discourse.
The idea of incorporating dynamic meaning representations into neural networks is not new. In the context of reading comprehension, \citet{Kobayashi16} proposed a model that dynamically computes the representation of a named entity mention from the local context given by its prior occurrences in the text.
In neural machine translation, the {\it copy mechanism} was proposed as a way of improving the handling of out-of-vocabulary words (e.g., named entities) in a source sentence~\citep{Gu16, Gulcehre16}.
We use a variant of recurrent neural language model (RNLM),
that combines dynamic representation and the copy mechanism.
The resulting novel model, \textit{Dynamic Neural Text Model}, uses
the dynamic word embeddings that are constructed from the context
in the output and input layers of an RNLM,
as shown in Figures~\ref{fig:dynamic_lm} and \ref{fig:context_merge}.

The contributions of this paper are three-fold.
First, we propose a novel neural language model, which we named the \textit{Dynamic Neural Text Model}.
Second, we introduce a new evaluation task and dataset called {\it Anonymized Language Modeling}. This dataset can be used to evaluate the ability of a language model to capture word meanings from contextual information (Figure~\ref{fig:example_calm}).
This task involves a kind of one-shot learning tasks,
in which the meanings of entities are inferred from their limited prior occurrences.
Third, our experimental results indicate that the proposed model
outperforms baseline models that use only global and static word embeddings
in the input and/or output layers of an RNLM.
Dynamic updates of the output layer
helps the RNLM predict reappearing entities,
whereas those of the input layer
are effective to predict words following reappearing entities.
A more detailed analysis showed that
the method was able to successfully capture the meanings of words across large contexts,
and to accumulate multiple context information.

\section{Background}
\label{sec:background}
\subsection{RNN Language Model}
\label{sec:RNNLM}

Given a sequence of $N$ tokens of a document $D=(w_1, w_2, ..., w_{N})$,
an RNN language model computes the probability
$p(D)=\prod_{t=1}^{N} p(w_t|w_1, ..., w_{t-1})$.
The computation of each factorized probability $p(w_t|w_1, ..., w_{t-1})$
can also be viewed as the task of predicting a following word $w_t$ from
the preceding words $(w_1, ..., w_{t-1})$.
Typically, RNNs recurrently compute the probability of the following word $w_{t}$ by using a hidden state $\bm{h}_{t-1}$ at time step $t-1$,
\begin{gather}
  p(w_t|w_1, ..., w_{t-1}) =
  \frac{ \exp({\vec{\bm{h}}_{t-1}}^{\intercal}\bm{y}_{w_t} + b_{w_t}) }{
    \sum_{w \in V} \exp({\vec{\bm{h}}_{t-1}}^{\intercal}\bm{y}_{w} + b_{w})} , \label{eq:softmax}\\
  \vec{\bm{h}}_{t} = \RNN (\bm{x}_{w_{t}}, \vec{\bm{h}}_{t-1}) .\label{eq:forward_def}
\end{gather}
Here, $\bm{x}_{w_{t}}$ and $\bm{y}_{w_{t}}$ denote the input and output word embeddings of $w_{t}$ respectively, $V$ represents the set of words in the vocabulary, and $b_{w}$ is a bias value applied when predicting the word $w$.
The function $\RNN$ is often replaced with LSTM~\citep{Hochreiter97} or GRU~\citep{Cho14} to improve performance.

\subsection{Dynamic Entity Representation}
\label{sec:der}
RNN-based models have been reported to achieve better results on the CNN QA reading comprehension dataset~\citep{Hermann15, Kobayashi16}.
In the CNN QA dataset, every named entity in each document is anonymized.
This is done to allow
the ability to comprehend a document
using neither prior nor external knowledge to be evaluated.
To capture the meanings of such anonymized entities,
\citet{Kobayashi16} proposed
a new model that they named {\it dynamic entity representation}.
This encodes the local contexts of an entity and
uses the resulting context vector as
the word embedding of a subsequent occurrence of that entity
in the input layer of the RNN.
This model:
(1) constructs context vectors $\bm{d}_{e,i}'$ from the local contexts of an entity $e$ at the $i$-th sentence;
(2) merges multiple contexts of the entity $e$ through max pooling and produces the dynamic representation $\bm{d}_{e,i}$;
and (3) replaces the embedding of the entity $e$ in the ($i+1$)-th sentence
with the dynamic embedding $\bm{x}_{e,i+1}$ produced from $\bm{d}_{e,i}$.
More formally,
\begin{align}
  \bm{x}_{e,i+1} & = W_{dc} \bm{d}_{e,i} + \bm{b}_{e} , \label{eq:context_vector_der} \\
  \bm{d}_{e,i} & = \mathrm{maxpooling}(\bm{d}_{e,i}', \bm{d}_{e,i-1}) , \label{eq:context_merger_max} \\
  \bm{d}_{e,i}' & = \mathrm{ContextEncoder}(e, i) . \label{eq:context_encoder}
\end{align}
Here, $\bm{b}_{e}$ denotes a bias vector,
$\mathrm{maxpooling}$ is a function that yields the largest value from the elementwise inputs,
and $\mathrm{ContextEncoder}$ is an encoding function.
Figure~\ref{fig:context_merge} gives an example of the process of encoding and merging contexts from sentences.
An arbitrary encoder can be used for $\mathrm{ContextEncoder}$;
\citet{Kobayashi16} used bidirectional RNNs, encoding the words surrounding the entity $e$ of a sentence in both directions.
If the entity $e$ fails to appear in the $i$-th sentence, the embedding is not updated, i.e., $\bm{d}_{e,i} = \bm{d}_{e,i-1}$.

\section{Proposed Method: Dynamic Neural Text Modeling}

In this section, we introduce the extension of dynamic entity representation to language modeling.
From Equations~\ref{eq:softmax} and \ref{eq:forward_def},
RNLM uses a set of word embeddings in the input layer to encode the preceding contextual words,
and another set of word embeddings in the output layer to predict a word from the encoded context.
Therefore, we consider incorporating the idea of dynamic representation
into the word embeddings in the output layer ($\bm{y}_w$ in Equation~\ref{eq:softmax})
as well as in the input layer ($\bm{x}_w$ in Equation~\ref{eq:forward_def}; refer to Figure~\ref{fig:dynamic_lm}).
The novel extension of dynamic representation to the output layer affects predictions made for entities that appear repeatedly,
whereas that in the input layer is expected to affect the prediction of words that follow the entities.

\begin{figure*}[!t]
  \begin{center}
    \includegraphics[width=13.5cm,clip]{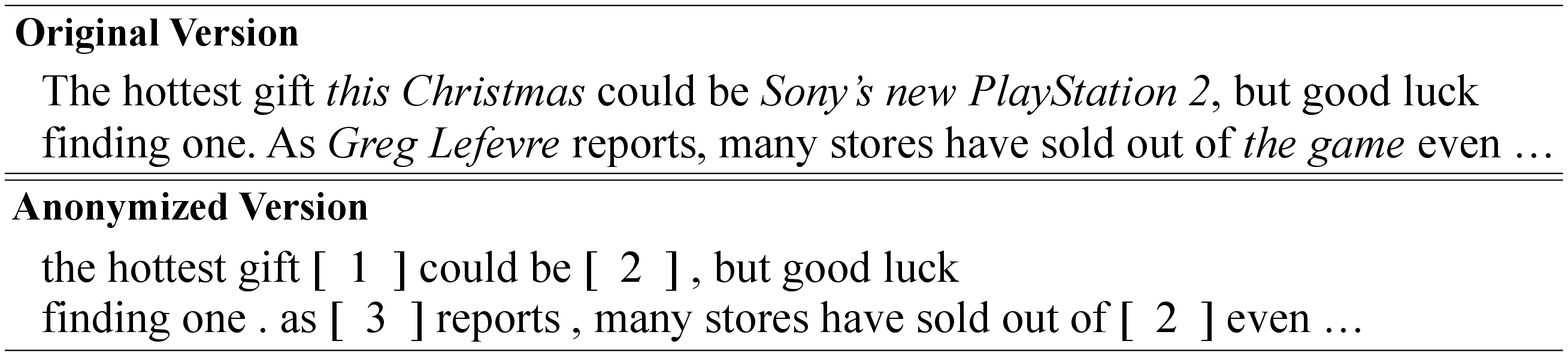}
    \caption{An example document for Anonymized Language Modeling.
      Token ``[ k ]'' is an anonymized token that appears k-th in the entities in a document.
      Language models predict the next word from the preceding words, and calculate probabilities for whole word sequences.}
    \label{fig:example_calm}
  \end{center}
\end{figure*}

The procedure for constructing dynamic representations of $e$, $\bm{d}_{e,i}$ is the same as that introduced in Section~\ref{sec:der}.
Before reading the ($i+1$)-th sentence,
the model constructs the context vectors $[\bm{d}_{e,1}', ..., \bm{d}_{e,i}']$
from the local contexts of $e$ in every preceding sentence.
Here, $\bm{d}_{e,j}'$ denotes the context vector of $e$ in the $j$-th sentence.
$\mathrm{ContextEncoder}$ in the model produces a context vector
$\bm{d}_{e}'$ for $e$ at the $t$-th position in a sentence,
using a bidirectional RNN\footnote{Equations \ref{eq:forward_def} and \ref{eq:forward_def2} are identical but do not share internal parameters.} as follows:
\begin{align}
  \bm{d}_{e}' & = \mathrm{ReLU}( W_{hd} [\vec{\bm{h}}_{t-1},\cev{\bm{h}}_{t+1} ] \!+\! \bm{b}_{d} ) ,\label{eq:enccontext_def} \\
  \vec{\bm{h}}_{t} & = \RNN (\bm{x}_{w_{t}}, \vec{\bm{h}}_{t-1}) \label{eq:forward_def2} ,\\
  \cev{\bm{h}}_{t} & = \RNNL (\bm{x}_{w_{t}}, \cev{\bm{h}}_{t+1}) \label{eq:backward_def} .
\end{align}
Here, $\mathrm{ReLU}$ denotes the ReLU activation function~\cite{Nair10},
while $W_{dc}$ and $W_{hd}$ correspond to learnable matrices;
$b_{d}$ is a bias vector.
As in the RNN language model,
$\vec{\bm{h}}_{t-1}$ and $\cev{\bm{h}}_{t+1}$ as well as their composition $\bm{d}_{e}'$
can capture information
necessary to predict the features of the target $e$ at the $t$-th word.

Following context encoding,
the model merges the multiple context vectors, $[\bm{d}_{e,1}', ..., \bm{d}_{e,i}']$,
into the dynamic representation $\bm{d}_{e,i}$ using a merging function.
A range of functions are abailable for merging multiple vectors,
while \citet{Kobayashi16} used only max pooling (Equation~\ref{eq:context_merger_max}).
In this study, we explored three further functions:
GRU, GRU followed by ReLU ($\bm{d}_{e,i} = \mathrm{ReLU}(\mathrm{GRU}(\bm{d}_{e,i}', \bm{d}_{e,i-1}))$)
and a function that selects only the latest context, i.e., $\bm{d}_{e,i} = \bm{d}_{e,i}'$.
This comparison clarifies the effect of the accumulation of contexts as the experiments proceeded\footnote{
Note that merging functions are not restricted to considering two arguments (a new context and a merged past context) recurrently
but can consider all vectors over the whole history $[\bm{d}_{e,1}', ..., \bm{d}_{e,i}']$ (e.g., by using attention mechanism~\citep{Bahdanau14}).
However, for simplicity, this research focuses only on the case of a function with two arguments.}.

The merging function produces the dynamic representation $\bm{d}_{e,i}$ of $e$.
In language modeling,
to read the $(i+1)$-th sentence,
the model uses two dynamic word embeddings of $e$
in the input and output layers.
The input embedding $\bm{x}_{e}$, used to encode contexts (Equation~\ref{eq:forward_def}),
and the output embedding $\bm{y}_{e}$, used to predict the occurrence of $e$ (Equation~\ref{eq:softmax}),
are replaced with dynamic versions:
\begin{align}
  \bm{x}_{e} & = W_{dx}\bm{d}_{e,i} + \bm{b}_{e}^x ,\label{eq:dx}\\
  \bm{y}_{e} & = W_{dy}\bm{d}_{e,i} + \bm{b}_{e}^y ,\label{eq:dy}
\end{align}
where $W_{dx}$ and $W_{dy}$ denote learnable matrices,
and $\bm{b}_{e}^x$ and $\bm{b}_{e}^y$ denote learnable vectors tied to $e$.
We can observe that a conventional RNN language model is a variant that
removes the dynamic terms ($W_{dx}\bm{d}_{e,i}$ and $W_{dy}\bm{d}_{e,i}$)
using only the static terms ($\bm{b}_{e}^x$ and $\bm{b}_{e}^y$)
to represent $e$.
The initial dynamic representation $\bm{d}_{e,0}$ is defined as a zero vector,
so that the initial word embeddings ($\bm{x}_{e}$ and $\bm{y}_{e}$) are identical to the static terms ($\bm{b}_{e}^x$ and $\bm{b}_{e}^y$)
until the point at which the first context of the target word $e$ is observed.
All parameters in the end-to-end model are learned entirely by
backpropagation, maximizing the log-likelihood in the same way as a conventional RNN language model.

We can view the approach in \citet{Kobayashi16} as a variant on the proposed method, but using the dynamic terms only in the input layer (for $\bm{x}_{e}$).
We can also view the copy mechanism~\citep{Gu16, Gulcehre16} as a variant on the proposed method, in which specific embeddings in the output layer are replaced
with special dynamic vectors.

\section{Anonymized Language Modeling}
\label{sec:calm}

This study explores
methods for on-the-fly capture and exploitation of the meanings of
unknown words or entities in a discourse.
To do this, we introduce a novel
evaluation task and dataset that we called
{\it Anonymized Language Modeling}.
Figure \ref{fig:example_calm} gives an example from the dataset.
Briefly, the dataset anonymizes
certain noun phrases, treating them as unknown words and retaining their coreference relations.
This allows
a language model to track the context of every noun phrase in the discourse.
Other words are left unchanged, allowing the language model
to preserve
the context of the anonymized (unknown) words, and to infer their meanings from the known words.
The process was inspired by \citet{Hermann15}, whose approach has been
explored by the research on reading comprehension.

More precisely,
we used the OntoNotes~\citep{Pradhan12} corpus,
which includes documents with coreferences and named entity tags manually annotated.
We assigned an anonymous identifier to every coreference chain in the corpus\footnote{We used documents with no more than 50 clusters, which covered more than 97\% of the corpus.} in order of first appearance\footnote{Following the study of \citet{Luong15rare}, we assigned ``\verb|<|unk1\verb|>|'', ``\verb|<|unk2\verb|>|'', ... to coreference clusters in order of first appearance.}, and replaced mentions of a coreference chain with its identifier.
In our experiments, each coreference chain was given a dynamic representation.
Following \citet{Mikolov10}, we limited the vocabulary to
10,000 words appearing frequently in the corpus.
Finally, we inserted ``\verb|<|bos\verb|>|'' and ``\verb|<|eos\verb|>|'' tokens
to mark the beginning and end of each sentence.

An important difference between this dataset and the one presented in \citet{Hermann15} is
in the way that coreferences are treated.
\citet{Hermann15} used automatic resolusion of coreferences, whereas our study made use of the manual annotations in the OntoNotes.
Thus, the process of \citet{Hermann15} introduced
(intentional and unintentional) errors into the dataset.
Additionally, the dataset did not assign an entity identifier to a pronoun.
In contrast, as our dataset has access to the manual annotations of coreferences, we are able to investigate the ability of the language model to capture meanings from contexts.

Dynamic updating could be applied to words in all lexical categories,
including verbs, adjectives, and nouns without requiring additional extensions.
However, verbs and adjectives
were excluded from targets of dynamic updates
in the experiments, for two reasons.
First, proper nouns and nouns accounted for the majority (70\%) of the low-frequency (unknown) words, followed by verbs (10\%) and adjectives (9\%).
Second, we assumed that the meaning of a verb or adjective would shift less over the course of a discourse than that of a noun.
When semantic information of unknown verbs and adjectives is required, their embeddings may be extracted from ad-hoc training on a different larger corpus.
This, however, was beyond the scope of this study.

\begin{table}
\small

\centering
\begin{tabular}{p{3.7cm}||ccc}
{\bf Split} & {\bf Train} & {\bf Valid} & {\bf Test} \\ \hline \hline
\# of documents & 2725 & 335 & 336 \\
Avg. \# of sentences & 25.7 & 27.2 & 26.4 \\
Avg. \# of unique entities & 15.6 & 16.8 & 15.8 \\
Avg. \# of unique entities occurring more than once & 9.3 & 9.9 & 9.5 \\
Avg. \# of occurrences of an entity & 3.2 & 3.2 & 3.1 \\
\end{tabular}
\caption{\label{tab:calm_stats} Statistics of Anonymized Language Modeling dataset.}
\end{table}

\section{Experiments}
\label{sec:experiment}

\setlength\tabcolsep{4.7pt}
\begin{table*}
\centering
\begin{tabular}{l||c|cc|c}
{\bf Models} &
{\bf (1) All} &
\shortstack{{\bf (2) Reappearing} \\ {\bf entities}} &
\shortstack{{\bf (3) Following} \\ {\bf entities}} &
{\bf (4) Non-entities}
\\ \hline \hline
LSTM LM (Baseline) (A) & 64.8$\pm$0.6 & 48.0$\pm$2.6 & 128.6$\pm$2.0 & 68.5$\pm$0.2\\
With only dynamic input (B) & 62.8$\pm$0.3 & 42.4$\pm$1.1 & 109.5$\pm$1.4 & \textbf{66.4$\pm$0.3}\\
With only dynamic output (C) & 62.5$\pm$0.3 & 35.9$\pm$3.7 & 129.0$\pm$0.7 & 69.5$\pm$0.3\\
With dynamic input \& output (D) & \textbf{60.7$\pm$0.2} & \textbf{34.0$\pm$1.3} & \textbf{106.8$\pm$0.6} & 67.6$\pm$0.04\\
\end{tabular}
\caption{\label{tab:results} Perplexities for each token group of models
  on the test set of Anonymized Language Modeling dataset.
  All values are averages with standard errors, calculated respectively by three models (trained with different random numbers).
  Dynamic models used GRU followed by ReLU as the merging function.}
\end{table*}

\subsection{Setting}
\label{sec:experimental_setting}

An experiment was conducted to investigate the effect of \textit{Dynamic Neural Text Model} on
the {\it Anonymized Language Modeling} dataset.
The split of dataset followed that of the original corpus~\citep{Pradhan12}.
Table \ref{tab:calm_stats} summarizes the statistics of the dataset.

The baseline model was a typical LSTM RNN language model with 512 units.
We compared three variants of the proposed model, using different applications of dynamic embedding:
in the input layer only (as in \citet{Kobayashi16}),
in the output layer only,
and in both the input and output layers.
The context encoders were bidirectional LSTMs with 512 units,
the parameters of which were not the same as those in the LSTM RNN language models.
All models were trained by maximizing the likelihood of correct tokens,
to achieve best perplexity on the validation dataset\footnote{
We performed a validation at the end of every half epoch out of five epochs.
}.
Most hyper-parameters
were tuned and fixed by the baseline model on the validation dataset\footnote{
Batchsize was 8.
Adam~\citep{Kingma15}
with learning rate $10^{-3}$.
Gradients were normalized so that their norm was smaller than 1.
Truncation of backpropagation
and updating was performed after every 20 sentences and at the end of document.
}.

It is difficult to adequately train the all parts of a model
using only the small dataset of Anonymized Language Modeling.
We therefore pretrained
word embeddings and $\mathrm{ContextEncoder}$ (the bi-directional RNNs and matrices in Equations \ref{eq:enccontext_def}--\ref{eq:backward_def}) on a sentence completion task
in which clozes were predicted from the surrounding words in a large corpus~\cite{Melanmug16}\footnote{
  We pretrained a model on the Gigaword Corpus, excluding
  sentences with more than 32 tokens.
  We performed training for 50000 iterations
  with a batch size of 128 and
  five negative samples.
  Only words that occurred no fewer than 500 times are used;
  other words were treated as unknown tokens.
  \citet{Melanmug16} used three different sets of word embeddings for the two inputs with respect to
  the encoders ($\RNN$ and $\RNNL$)
  and the output (target).
  However, we forced the sets of word embeddings to share a single set of word embeddings in pretraining.
  We initialized the word embeddings in both the input layer ($\bm{x}_w$) and the output layer ($\bm{y}_w$)
  of the novel models, including the baseline model, with this single set.
  The word embeddings of all anonymized tokens were initialized as unknown words with the word embedding of ``\verb|<|unk\verb|>|''.
}.
We used the objective function with negative sampling~\citep{Mikolov13nips}:
$\sum_e (\log\sigma({\hat{\bm{x}}_{e}}^{\intercal} {\bm{x}_{e}}) +
\sum_{v\in Neg}(\log\sigma( -{\hat{\bm{x}}_{e}}^{\intercal} {\bm{x}_{v}} )))$.
Here, $\hat{\bm{x}}_e$ is a context vector predicted by $\mathrm{ContextEncoder}$,
$\bm{x}_e$ denotes the word embedding of a target word $e$ appearing in the corpus,
and $Neg$ represents randomly sampled words.
These pretrained parameters of $\mathrm{ContextEncoder}$
were fixed when the whole language model was trained on the Anonymized Language Modeling dataset.
We implemented models in Python using the {\tt Chainer} neural network library~\citep{Tokui15}.
The code and the constructed dataset are publicly available\footnote{\url{https://github.com/soskek/dynamic_neural_text_model}}.

\subsection{Results and Analysis}

\subsubsection{Perplexity}

Table~\ref{tab:results} shows performance of the baseline model and the three variants of the proposed method in terms of perplexity.
The table reports the mean and standard error of three perplexity values after training using three different randomly chosen initializations (we used the same convention throughout this paper).
Here, we discuss the proposed method using GRU followed by ReLU as the merging function, as this achieved the best perplexity (see Section~\ref{sec:comp_merge} for a comparison of functions).
We also show perplexitiy values when evaluating words of specific categories: (1) all words; (2) reappearing entity words; (3) words following entities; and (4) non-entity words.

All variants of the proposed method outperformed the baseline model.
Focusing on the categories (2) and (3) highlights the roles of dynamic updates of the input and output layers.
Dynamic updates of the input layer (B) had a larger improvement for predicting words following entities (3) than those of the output layer (C).
In contrast, dynamic updates of the output layer (C) were quite effective for predicting reappearing entities (2) whereas those of the input layer (B) were not.
These facts confirm that: dynamic updates of the input layer help a model predict words following entities by supplying on-the-fly context information;
and those of the output layer are effective to predict entity words appearing multiple times.

In addition, dynamic updates of both the input and output layers (D) further improved the performance from those of either the output (C) or input (B) layer.
Thus, the proposed dynamic output was shown to be compatible with dynamic input, and {\it vice versa}.
These results demonstrated the positive effect of capturing and exploiting the context-sensitive meanings of entities.

\begin{figure}[t]
  \begin{center}
    \includegraphics[width=6.8cm,clip]{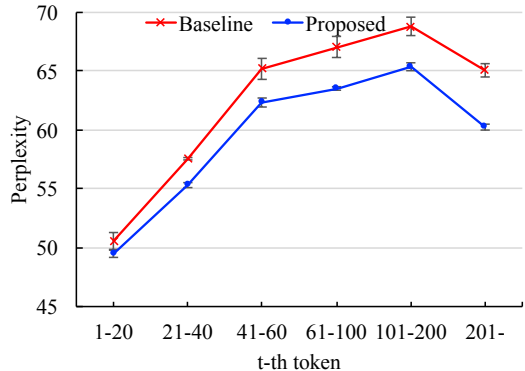}
    \caption{Perplexity of all tokens relative to the time at which they appear in the document.}
    \label{fig:ith_token}
  \end{center}
\end{figure}

\begin{figure}[t]
  \begin{center}
    \includegraphics[width=6.8cm,clip]{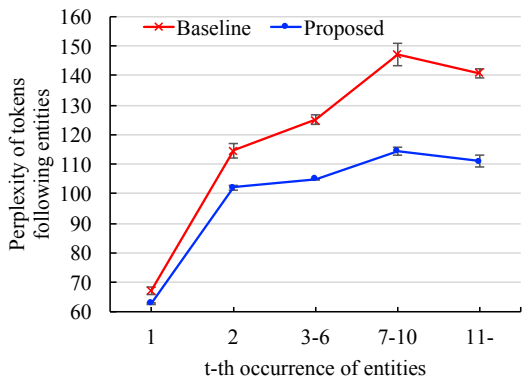}
    \caption{Perplexity of tokens following the entities relative to the time at which the entity occurs.}
    \label{fig:ith_occurence_after}
  \end{center}
\end{figure}

\begin{figure}
  \begin{center}
    \includegraphics[width=6.8cm,clip]{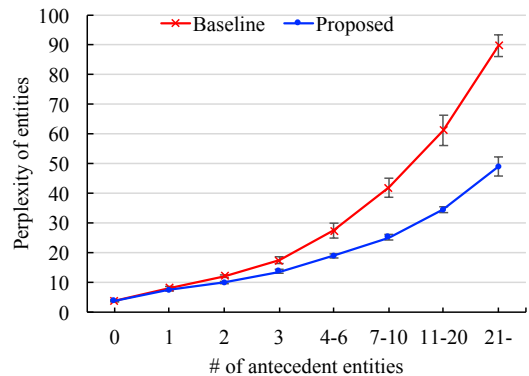}
    \caption{Perplexity of entities relative to the number of antecedent entities.}
    \label{fig:n_ants_entity}
  \end{center}
\end{figure}

In order to examine whether dynamic updates of the input and output embeddings capture context-sensitive meanings of entities, we present Figures~\ref{fig:ith_token}, \ref{fig:ith_occurence_after} and \ref{fig:n_ants_entity}.
Figure~\ref{fig:ith_token} depicts the perplexity of words with different positions in a document\footnote{
It is more difficult to predict tokens appearing latter in a document because the number of new (unknown) tokens increases as a model reads the document.}.
The figure confirms that the advantage of the proposed method over the baseline is more evident especially in the latter part of documents, where repeated words are more likely to occur.

Figure~\ref{fig:ith_occurence_after} shows the perplexity with respect to the frequency of words $t$ within documents.
Note that the word embedding at the first occurrence of an entity is static.
This figure indicates that entities appearing many times enjoy the benefit of the dynamic language model.
Figure~\ref{fig:n_ants_entity} visualizes the perplexity of entities with respect to the numbers of their antecedent candidates.
It is clear from this figure that the proposed method is better at memorizing the semantic information of entities appearing repeatedly in documents than the baseline.
These results also demonstrated the contribution of dynamic updates of word embeddings.


\setlength\tabcolsep{4.2pt}
\begin{table*}
\small
\centering
\begin{tabular}{lll||c|cc|c}
{\bf Models} & {\bf Merging function} &
\shortstack{{\bf \# of parameters} \\ {\bf (to be finetuned)}} &
{\bf (1) All} &
\shortstack{{\bf (2) Reappearing} \\ {\bf entities}} &
\shortstack{{\bf (3) Following} \\ {\bf entities}} &
{\bf (4) Non-entities}
\\ \hline \hline
Only & GRU-ReLU & 18.9M (14.2M) & \textbf{62.8$\pm$0.3} & \textbf{42.4$\pm$1.1} & \textbf{109.5$\pm$1.4} & \textbf{66.4$\pm$0.3}\\
dynamic input & GRU & 18.9M (14.2M) & 63.2$\pm$0.4 & 43.3$\pm$2.7 & 111.2$\pm$0.7 & 66.8$\pm$0.4\\
 & Max pool. & 17.3M (12.6M) & 63.6$\pm$0.4 & 45.0$\pm$2.6 & 116.0$\pm$1.0 & 67.0$\pm$0.2\\
 & Only latest & 17.3M (12.6M) & 64.0$\pm$0.4 & 44.1$\pm$1.6 & 127.6$\pm$0.7 & 67.5$\pm$0.2\\

\hline
Only & GRU-ReLU & 18.9M (14.2M) & 62.5$\pm$0.3 & \textbf{35.9$\pm$3.7} & 129.0$\pm$0.7 & 69.5$\pm$0.3\\
dynamic output & GRU & 18.9M (14.2M) & 62.6$\pm$0.2 & 39.0$\pm$2.0 & \textbf{121.1$\pm$8.3} & 69.1$\pm$0.2\\
 & Max pool. & 17.3M (12.6M) & \textbf{62.2$\pm$0.4} & 41.1$\pm$1.9 & 126.9$\pm$1.5 & \textbf{68.4$\pm$0.6}\\
 & Only latest & 17.3M (12.6M) & 64.9$\pm$0.1 & 49.8$\pm$1.8 & 129.1$\pm$1.6 & 70.6$\pm$0.2\\

\hline
Dynamic & GRU-ReLU & 19.2M (14.4M) & \textbf{60.7$\pm$0.2} & \textbf{34.0$\pm$1.3} & \textbf{106.8$\pm$0.6} & 67.6$\pm$0.04\\
input \& output & GRU & 19.2M (14.4M) & 60.9$\pm$0.3 & 37.5$\pm$0.3 & 108.9$\pm$0.8 & 67.2$\pm$0.4\\
 & Max pool. & 17.6M (12.9M) & \textbf{60.7$\pm$0.3} & 39.5$\pm$3.4 & 107.5$\pm$1.3 & \textbf{66.8$\pm$0.8}\\
 & Only latest & 17.6M (12.9M) & 63.4$\pm$0.2 & 47.9$\pm$4.2 & 116.4$\pm$0.4 & 68.9$\pm$0.1\\
\hline
\multicolumn{2}{l}{Baseline} & 12.3M (12.3M) & 64.8$\pm$0.6 & 48.0$\pm$2.6 & 128.6$\pm$2.0 & 68.5$\pm$0.2\\

\end{tabular}
\caption{\label{tab:results_merge} Results for models with different merging functions on the test set of the Anonymized Language Modeling dataset, as same as in Table 2.}
\end{table*}

\subsubsection{Comparison of Merging functions}
\label{sec:comp_merge}

Table~\ref{tab:results_merge} compares models
with different merging functions; GRU-ReLU, GRU, max pooling, and the use of the latest context.
The use of the latest context had the worst performance for all variants of the proposed method.
Thus, a proper accumulation of multiple contexts is indispensable for dynamic updates of word embeddings.
Although \citet{Kobayashi16} used only max pooling as the merging function,
GRU and GRU-ReLU were shown to be
comparable in performance and superior to max pooling
when predicting tokens related to entities (2) and (3).

\subsubsection{Predicting Entities by Likelihood of a Sentence}

In order to examine contribution of the dynamic language models on a downstream task, we conducted cloze tests for comprehension of a sentence with reappearing entities in a discourse.
Given multiple preceding entities $E = \{e^{+}, e^{1}, e^{2}, ...\}$ followed by a cloze sentence, the models were required to predict the true antecedent $e^{+}$ which allowed the cloze to be correctly filled, among the other alternatives $E^{-} = \{e^{1}, e^{2}, ...\}$.

Language models solve this task
by comparing the likelihoods of sentences filled with antecedent candidates in $E$
and returning the entity with the highest likelihood of the sentence.
In this experiment,
the performance of a model was represented by
the {\it Mean Quantile} (MQ)~\citep{Guu15}.
The MQ computes the mean ratio at which
the model predicts a correct antecedent $e^{+}$ more likely
than negative antecedents in $E^{-}$,
\begin{align}
  \mathrm{MQ} & = \frac{|\{e^{-} \in E^{-}: p(e^{-}) < p(e^{+})\}|}{|E^{-}|} . \label{eq:mq}
\end{align}
Here, $p(e)$ denotes the likelihood of a sentence whose cloze is filled with $e$.
If the correct antecedent $e^{+}$ yields highest likelihood,
MQ gets 1.

Table~\ref{tab:mqs} reports MQs for the three variants and merging functions.
Dynamic updates of the input layer greatly boosted the performance by approximately 10\%,
while using both dynamic input and output improved it further.
In this experiment, the merging functions with GRUs outperform the others.
These results demonstrated that Dynamic Neural Text Models can accumulate a new information in word embeddings and contribute to modeling the semantic changes of entities in a discourse.

\begin{table}
\small
\centering
\begin{tabular}{ll||c}
{\bf Models} & {\bf Merging func.} & {\bf MQ} \\ \hline \hline
Baseline & & .525$\pm$.001 \\ \hline
Only & GRU-ReLU & .630$\pm$.005 \\
dynamic input & GRU & .633$\pm$.005 \\
 & Max pool. & .617$\pm$.002 \\
 & Only latest & .600$\pm$.004 \\ \hline
Only & GRU-ReLU & .519$\pm$.001 \\
dynamic output & GRU & .522$\pm$.000 \\
 & Max pool. & .519$\pm$.001 \\
 & Only latest & .519$\pm$.003 \\ \hline
Dynamic & GRU-ReLU & \textbf{.642$\pm$.004} \\
input \& output & GRU & .637$\pm$.005 \\
 & Max pool. & .620$\pm$.002 \\
 & Only latest & .613$\pm$.002 \\
\end{tabular}
\caption{\label{tab:mqs}
  Mean Quantile of a true coreferent entity among antecedent entities.}
\end{table}

\section{Related Work}
\label{sec:related_work}

An approach to addressing the unknown word problem
used in recent studies~\citep{Kim16,Sennrich16,Luong16,Schuster12}
comprises the embeddings of unknown words from
character embeddings or subword embeddings.
\citet{Li15} applied word disambiguation and use a sense embedding to the target word.
\citet{Choi17}
captured the context-sensitive meanings of common words
using word embeddings, applied through a gating function controlled by history words,
in the context of machine translation.
In future work,
we will explore a wider range of models,
to integrate our dynamic text modeling with methods that estimate the meaning of unknown words or entities from their constituents.
When addressing well-known entities such as \textit{Obama} and \textit{Trump}, it makes sense to learn their embeddings from external resources, as well as dynamically from the preceding context in a given discourse (as in our Dynamic Neural Text Model). The integration of these two sources of information is an intriguing challenge in language modeling.

A key aspect of our model is its incorporation of the copy mechanism~\citep{Gu16, Gulcehre16}, using dynamic word embeddings in the output layer.
Independently of this study,
several research groups have explored the use of variants of the copy mechanisms
in language modeling~\citep{Merity16,Grave16,Peng16}.
These studies, however, did not incorporate
dynamic representations in the input layer.
In contrast, our proposal incorporates the copy mechanism through the use of dynamic representations in the output layer,
integrating them with dynamic mechanisms in both the input and output layers by applying dynamic entity-wise representation. Our experiments have demonstrated the benefits of such integration.

Another related trend in recent studies is the use of neural network to capture the information flow of a discourse.
One approach has been to link RNNs across sentences~\citep{Wang16,Serban16},
while a second approach has expolited a type of memory space to store
contextual information~\citep{Sukhbaatar15,Tran16,Merity16}.
Research on reading comprehension~\citep{Kobayashi16,Henaff16} and coreference resolution~\citep{Wiseman16,Clark16acl,Clark16emnlp}
has shown the salience of
entity-wise context information.
Our model could be located within such approaches,
but is distinct in being the first model
to make use of entity-wise context information
in both the input and output layers for sentence generation.

We summarize and compare works for entity-centric neural networks that read a document.
\citet{Kobayashi16} pioneered entity-centric neural models tracking states in a discourse. They proposed {\it Dynamic Entity Representation}, which encodes contexts of entities and updates the states using entity-wise memories. \citet{Wiseman16} also proposed a method for managing similar entity-wise features on neural networks and improved a coreference resolution model. \citet{Clark16acl,Clark16emnlp} incorporated such entity-wise representations in mention-ranking coreference models.
Our paper follows \citet{Kobayashi16} and exploits dynamic entity reprensetions in a neural language model, where dynamic reporesentations are used not only in the neural encoder but also in the decoder, applicable to various sequence generation tasks, e.g., machine translation and dialog response generation. Simultaneously with our paper, \citet{Ji17} use dynamic entity representation in a neural language model for reranking outputs of a coreference resolution system. \citet{Yang17} experiment language modeling with referring to internal contexts or external data. \citet{Henaff16} focus on neural networks tracking contexts of entities, achieving the state-of-the-art result in bAbI~\citep{Weston15}, a reading comprehension task. They encode the contexts of each entity by an attention-like gated RNN instead of using coreference links directly. \citet{Dhingra17} also try to improve a reading comprehension model using coreference links. Similarly to our dynamic entity representation, \citet{Bahdanau17} construct on-the-fly word embeddings of rare words from dictionary definitions.

The first key component of dynamic entity representation is a function to merge more than one contexts about an entity into a consistent representation of the entity. Various choices for the function exist, e.g., max or average-pooling~\citep{Kobayashi16,Clark16acl}, RNN (GRU, LSTM~\citep{Wiseman16,Yang17} or other gated RNNs~\citep{Henaff16,Ji17}), or using the latest context only (without any merging)~\citep{Yang17}. This paper is the first work comparing the effects of those choices (see Section~\ref{sec:comp_merge}).

The second component is a function to encode local contexts from a given text, e.g., bidirectional RNN encoding~\citep{Kobayashi16}, unidirectional RNN used in a language model~\citep{Ji17,Yang17}, feedforward neural network with a sentence vector and an entity's word vector~\citep{Henaff16} or hand-crafted features with word embeddings~\citep{Wiseman16,Clark16acl}. This study employs bi-RNN analogously to \citet{Kobayashi16}, which can access full context with powerful learnable units.

In the task setting proposed in this study,
a model must capture the meaning of a given specific word from a small number of its contexts in a given discourse.
The task could also be seen as novel one-shot learning~\citep{Fei06} of word meanings.
One-shot learning for NLP like this has been little studied, with the exception of the study by \citet{Vinyals16},
which used a task in which
the context of a target word is matched with a different context of the same word.

\section{Conclusion}
This study addressed the
problem of identifying the meaning of unknown words or entities in a discourse
with respect to the word embedding approaches used in
neural language models.
We proposed a method for on-the-fly construction and exploitation of
word embeddings in both the input layer and output layer
of a neural model by tracking contexts.
This extended the dynamic entity representation presented in \citet{Kobayashi16},
and incorporated a copy mechanism proposed independently by \citet{Gu16} and \citet{Gulcehre16}.
In the course of the study, we also constructed
a new task and dataset, called
{\it Anonymized Language Modeling},
for evaluating the ability of a model to capture word meanings while reading.
Experiments conducted using our novel dataset
demonstrated that the RNN language model variants proposed in this study
outperformed the baseline model.
More detailed analysis
indicated that the proposed method was
particularly
successful in capturing
the meaning of an unknown words from texts containing few instances.

\section*{Acknowledgments}

This work was supported by JSPS KAKENHI Grant Number 15H01702 and JSPS KAKENHI Grant Number 15H05318.
We thank members of Preferred Networks, Inc., Makoto Miwa and Daichi Mochihashi for suggestive discussions.

\bibliography{ijcnlp2017}
\bibliographystyle{ijcnlp2017}

\end{document}